\documentclass[letterpaper, 10 pt, conference]{ieeeconf}  

\IEEEoverridecommandlockouts                              

\usepackage{graphics} 
\usepackage{graphicx}
\usepackage{amsmath}
\usepackage{amssymb}
\usepackage{booktabs}
\usepackage{multicol}
\usepackage{multirow}
\usepackage{url}
\usepackage{caption}
\usepackage{subcaption}
\usepackage{setspace}
\usepackage{xcolor}
\usepackage{cleveref}
\definecolor{myred}{HTML}{ff0000}
\definecolor{mypurple}{HTML}{8020cb}  
\definecolor{myorange}{HTML}{F0A73A}

\title{\LARGE \bf
GenTe: Generative Real-world Terrains for General Legged Robot Locomotion Control
}

\author{Hanwen Wan$^{1, 2}$, Mengkang Li$^{1, 2}$, Donghao Wu$^{1, 2}$, Yebin Zhong$^{1, 2}$, Yixuan Deng$^{1, 2}$\\ Zhenglong Sun$^{1, 2}$, and Xiaoqiang Ji$^{1, 2, \dagger}$
\thanks{$^{1}$School of Science and Engineering, The Chinese University of Hong Kong, Shenzhen, China.}%
\thanks{$^{2}$Shenzhen Institute of Artificial Intelligence and Robotics for Society, China.}%
\thanks{$^{\dagger}$The corresponding author is Xiaoqiang Ji whose e-mail is {\tt\small jixiaoqiang@cuhk.edu.cn}}%
}

\begin{document}

\maketitle
\thispagestyle{empty}
\pagestyle{empty}

\begin{abstract}
Developing bipedal robots capable of traversing diverse real-world terrains presents a fundamental robotics challenge, as existing methods using predefined height maps and static environments fail to address the complexity of unstructured landscapes. To bridge this gap, we propose GenTe, a framework for generating physically realistic and adaptable terrains to train generalizable locomotion policies. GenTe constructs an atomic terrain library that includes both geometric and physical terrains, enabling curriculum training for reinforcement learning-based locomotion policies. By leveraging function-calling techniques and reasoning capabilities of Vision-Language Models (VLMs), GenTe generates complex, contextually relevant terrains from textual and graphical inputs. The framework introduces realistic force modeling for terrain interactions, capturing effects such as soil sinkage and hydrodynamic resistance. To the best of our knowledge, GenTe is the first framework that systemically generates simulation environments for legged robot locomotion control. Additionally, we introduce a benchmark of 100 generated terrains. Experiments demonstrate improved generalization and robustness in bipedal robot locomotion.

\end{abstract}


\section{INTRODUCTION}

Bipedal robot locomotion control remains a multifaceted challenge in robotics, with terrain robustness and generalization representing critical unsolved capabilities \cite{c1, c2}. Current methods primarily train agents in simulation via reinforcement learning \cite{c3}, operating on two distinct terrain abstractions: (1) geometric terrains, modeled as rigid surfaces parameterized by height maps; and (2) physical terrains, which introduce dynamic force interactions through ground compliance or fluid dynamics.

\begin{figure}[t]
    \centering
    \includegraphics[width=1\linewidth]{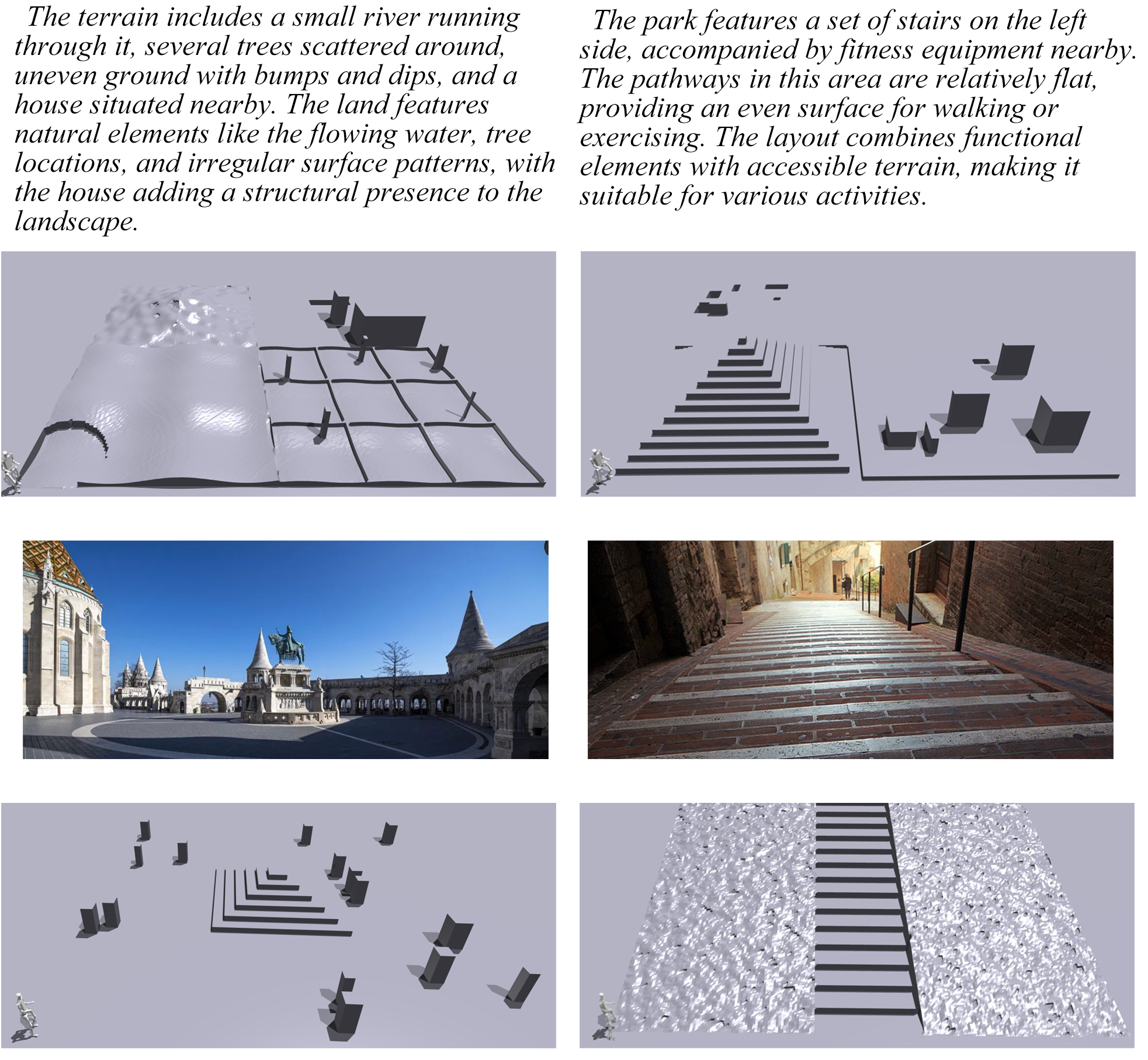}
    \caption{Terrains generated by GenTe with corresponding text/image prompts.}
    \label{fig:1}
\end{figure}

Prior methods for training legged robots to generalize across terrains typically use rigid, pre-defined environments by adjusting height maps to simulate standard surfaces like flat ground, stairs, and slopes \cite{c4}. While these approaches establish basic locomotion capabilities, they fall short of capturing the shifting dynamics of natural terrains \cite{c5}. Such complex and hybrid terrains require adaptive behaviors in response to unpredictable conditions. Some works focus on enhancing robustness by precisely modeling the specialized forces involved, such as sand \cite{c6} and snow \cite{c7}. Integrating terrain modeling into learning-based locomotion control for legged robots facilitates the modeling of complex terrains that exceed the capabilities of conventional simulators \cite{c9, c10}. However, current simulators have limitations in accurately simulating the nuanced physical properties of real-world terrains. Although domain randomization \cite{c8} introduces minor variations in texture and height, this alone fails to replicate the nuanced physical interactions found in natural terrains, limiting the adaptability of learned policies in diverse and unstructured environments.

To address this gap, we present \textbf{GenTe}, a comprehensive framework for \textbf{Gen}erating real-world \textbf{Te}rrains to train legged robot policies. GenTe systematically constructs atomic terrain components into a versatile library, constructs realistic terrains leveraging the reasoning capabilities of Vision-Language Models (VLMs). Specifically, geometric terrains are represented through height maps, while physical terrains are designed using force simulations. We account for additional forces resulting from terrain characteristics, such as soil sinkage and hydrodynamic drag, and apply them to the robot's movement. To further enhance the realism of terrain generation, we introduce a novel approach that leverages function-calling techniques. Each basic terrain type is formulated as a tool, with specific parameters that define its physical properties and behavioral characteristics. For example, a beach environment includes geometric features like flat ground with sporadic rocky outcrops, along with physical characteristics such as wading and deformable sand. By calling specific functions within the terrain generation framework, we can create terrains with tailored physical attributes. GenTe is capable of transforming both figures and textual descriptions into contextually relevant terrains for simulation. Based on this framework, we also introduce a terrain benchmark containing 100 generated terrains—50 generated from text and 50 from figures. These terrains have been reviewed by human evaluators to ensure quality and accuracy. The code and benchmark are open-sourced\footnote{\url{https://github.com/HaronW/GenTe}}. The main contributions of this work are as follows:

\begin{itemize}
    \item  \textbf{Development of a terrain library} for legged robots, incorporating both geometric and physical features to cover a wide range of real-world surface conditions, enabling bipedal robots to adapt to diverse surfaces including deformable and fluid-based terrains.
    \item \textbf{Terrain generation framework GenTe} allows flexible and scalable terrain creation based on textual or graphical input. To the best of our knowledge, GenTe is the first framework for generalized locomotion simulation environment generation.
    \item \textbf{Comprehensive experiments} are conducted to evaluate the proposed methods, demonstrating the effectiveness of the RL-trained robot in traversing different terrains and the LLM’s ability.
\end{itemize}

\section{RELATED WORKS}

\subsection{Terrain Modeling for Robot Locomotion}
Terrain modeling relies on dynamic and mechanical principles to ensure precision in simulations. For deformable terrains like sand and snow, early efforts \cite{c11,c12} employed the Bekker pressure-sinkage relation and the Janosi-Hanamoto shear-displacement equation to model contact shear forces. Meanwhile, modern research often proposes mathematical models that align with experimental results \cite{c6, c13}. For example, \cite{c15} focuses on estimating terramechanical properties of deformable terrains and introduces a practical model for lightweight legged robots. This model simplifies foot-terrain interaction into two dominant forces: gross sliding friction and bulldozing resistance.

In parallel, fluid mechanics and hydrodynamics have been utilized to simulate fluid-terrain interactions, enabling robots to navigate aquatic environments \cite{c15, c16}. Gazebo Fluids \cite{c17} extends the Gazebo simulator to model fluid interactions with articulated robots through smoothed particle hydrodynamics. This approach captures fluid dynamics around complex robot structures, simulating forces such as drag, buoyancy, and added mass. These advancements provide a foundation for more sophisticated simulation techniques that improve robot adaptability across diverse terrains. However, despite the significant progress made in terrain modeling, few studies have integrated various types of terrains into a single comprehensive framework.

\subsection{Simulation Environment Generation for Embodied Agents}
Current efforts on simulation environment generation have predominantly focused on manipulation tasks \cite{c18} with flexible layouts and object placement. For instance, MimicGen \cite{c19} creates highly varied virtual environments, allowing agents to learn robust skills across different settings without relying on real-world data. This approach facilitates generalization across a broad range of scenarios, enhancing the agents' adaptability to real-world applications. DexMimicGen \cite{c20} builds on this by simulating complex interactions with objects, particularly emphasizing fine-grained control of robotic hands and fingers, thereby improving dexterous manipulation in diverse environments. ProcTHOR \cite{a1} generates simulation environments using procedural generation techniques, which allow for the automatic creation of diverse, 3D environments. This approach allows agents to learn robust navigation and interaction policies across a wide range of environments, from simple rooms to intricate layouts.

However, simulation generation for locomotion has not received as much attention. Simulators focus on generating environments for basic movement tasks but often lack the complexity required for simulating dynamic, unpredictable terrains or interactions with multiple environmental factors

\section{TERRAIN SIMULATION}
Contemporary robotic simulation platforms often fall short in accurately representing complex terrains like beaches, sand, or snow, which have unique physical properties critical for bipedal robot training. These terrains require simulating intricate forces such as terrain deformation, friction, and resistance, necessitating the integration of terramechanics principles to model robot-terrain interactions. Existing simulators fail to adequately capture these dynamics, highlighting the need for custom terrain models. Our framework addresses this by combining two complementary levels: the geometry level, which uses height maps to model terrain features like hills and obstacles, and the physics level, which incorporates force simulations based on terramechanics to model soil deformation, friction, and hydrodynamic drag. This dual-level approach enables the creation of realistic and diverse training terrains, better preparing robots for real-world deployment. Examples are shown in \cref{fig:1}.


\subsection{Geometry Terrains}
At the geometry level, we construct terrain surfaces using a combination of height maps and geometric primitives. Height maps provide a versatile way to define the underlying topography, allowing for the creation of diverse features such as hills, valleys, and uneven ground. These height maps can be generated procedurally using mathematical functions or by leveraging data from real-world terrain scans. 

In addition to the height information, we also incorporate geometric shapes and objects to add further complexity to the terrain. These can include obstacles like rocks and trees, as well as transitional elements like slopes and cliffs. By blending height maps with strategic placement of these geometric features, we can generate a wide range of terrain types, from smooth, rolling landscapes to more rugged, obstacle-laden environments.

\subsection{Physical Terrain}
\subsubsection{Wading Terrain}
\begin{figure}[t]
  \centering
  \includegraphics[width=0.95\linewidth]{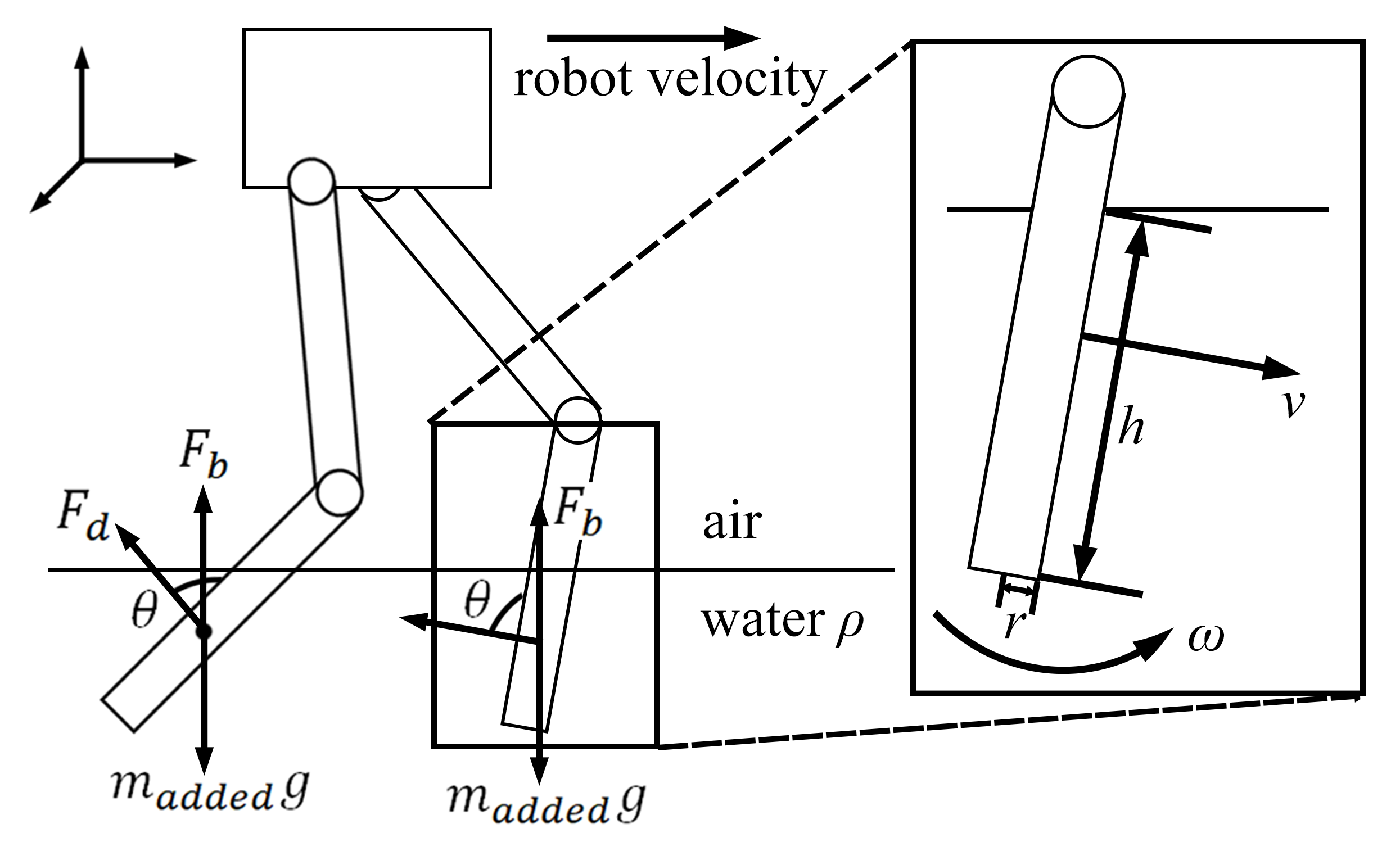}
  \caption{Analysis of bipedal robots walking on wading terrains.}
  \label{fig:wading}
\end{figure}

The wading terrain simulates the challenges of locomotion in water, where the bipedal robot encounters forces such as fluid resistance, added mass effects, and buoyancy, as shown in \cref{fig:wading}. To replicate real-world aquatic conditions, an additional flow-induced force, is introduced to model the influence of water movement. This force is represented through sinusoidal, Gaussian, and zero-flow patterns, corresponding to dynamic conditions such as ocean waves, streams, and still water, respectively. These variations enable the robot to experience and adapt to a wide range of fluid forces, closely approximating the complexities found in natural water environments.

The forces are decomposed into horizontal and vertical components. In the simulations, the horizontal forces are converted into torques acting on the robot's center of mass, while the total vertical force is reflected by adjusting the robot's effective weight. To enhance generalization, Gaussian noise $\epsilon \sim \mathcal{N}(1, 0.1)$ is applied to both the horizontal and vertical force components. This noise parameter is initialized at the start of each simulation period and remains constant throughout that period.

The \textbf{drag force} on the robot's legs consists of two components: form drag, resulting from pressure differences across the surface, and viscous drag, arising from shear stress along the leg's surface due to the boundary layer. The total drag force $F_d$ acting on a leg can be expressed as:

\begin{equation}
    F_d(t) = \frac{1}{2} \epsilon C_d \rho A(t) v(t)^2
    \label{eq:drag}
\end{equation}
where $\rho$ is the water density, $C_d$ is the drag coefficient, and $v(t)$ is the velocity of the leg through water. $A(t)$ is the projected area of the leg perpendicular to the flow, which can be computed by $A(t) = 2 \pi r h(t)$, where $r$ is leg radius, $h$ is submerged leg length. The drag coefficient $C_d$ depends on the Reynolds number $Re$, as described in:

\begin{equation}
    \begin{split}
        Re = \frac{\rho vL}{\mu} = \frac{1025 \times 1 \times 0.1}{0.0011} \approx 93182, 
    \end{split}
    \label{eq:Re}
\end{equation}
where $L$ is the characteristic length, approximated by the leg's diameter $2r = 0.1$ meters, $\mu$ is the dynamic viscosity of the water, approximated as $\mu = 0.0011 Pa \cdot s$. With $Re$ exceeding 4000, the flow regime is turbulent, and $C_d$ remains approximately constant for cylindrical shapes, typically ranging from 0.82 to 1.0.

Applying the drag force $F_d$ to the robot is implemented using simulator APIs, allowing for real-time computation of the hydrodynamic forces based on the robot's movement through the water. The drag force is calculated for each leg independently, as \cref{eq:drag} considers only one leg at a time. Since the submerged length $h(t)$, projected area $A(t) = 2 \pi r h(t)$, and velocity $v(t)$ vary over time and differ between the two legs, the drag forces acting on each leg are unique and time-dependent, reflecting the dynamic nature of their movement.

The \textbf{added mass effect} occurs because the leg accelerates not only itself but also the surrounding water, increasing the system's effective inertia and resulting in addition power. The added mass $m_\text{added}$ is given by:

\begin{equation}
    m_\text{added}(t) = C_m \rho V_\text{sub}(t),
\end{equation}
where $C_m$ is the added mass coefficient (approximately ~0.5 for cylindrical shapes) and $V_\text{sub}(t) = \pi r^2 h(t)$ is the submerged volume of the leg.

The \textbf{buoyant force} $F_b$ is determined by the submerged volume $V_\text{sub}$ of the robot’s legs and the water’s pressure gradient, corresponding to the weight of the displaced fluid. To simplify the simulation of vertical forces, the buoyant effect is incorporated by adjusting the robot’s effective mass, while assuming a constant water density $\rho$. This approach allows for efficient computation by treating the buoyant force as a reduction in the robot’s perceived weight as it moves through water. Considering both legs, the robot's effective weight $m_\text{eff}$ is expressed as:

\begin{equation}
    \begin{split}
        m_\text{eff}(t) &= \epsilon \cdot \frac{G + m_\text{added}(t)g - F_b(t)}{g} \\
        &= \epsilon \left(m + C_m \rho V_\text{sub, left}(t) + C_m \rho V_\text{sub, right}(t) - \rho V_\text{sub, left}(t) - \rho V_\text{sub, right}(t)) \right),
    \end{split}
\end{equation}
where $G = mg$ is the robot’s gravity. In this form, the effective mass depends on the depth of submersion $h$ and the fluid's properties. The effective mass can be expressed as a function of the time-dependent depth $h(t)$. This equation accounts for variations in the leg’s submersion as the robot moves through water, thereby dynamically updating the effective mass based on the changing fluid forces.

The \textbf{water flow force} simulate constant water, currents and tide. Additionally, the water flow force is perturbed by a dynamic Gaussian noise term $\xi(t) \sim \mathcal{N}(1, 0.1)$, accounting for environmental variability. This noise term modifies the overall force experienced by the robot over time. The water flow force $F_\text{flow}(t)$ is calculated as follows:

\begin{equation}
    F_\text{flow}(t) = 
        \begin{cases}
        0 & \text{constant water}, \\
        F_\text{current} & \text{current}, \\
        F_\text{tide} cos(\omega t+ \phi) & \text{tide},
        \end{cases}
\end{equation}
where $F_\text{current}$ and $F_\text{tide}$ are the amplitude, $\omega$ is the angular frequency of the tide, and $\phi$ is the phase shift.

In the simulation of wading terrain, key parameters such as leg dimensions, water density, and fluid properties are integrated into the equations governing the forces acting on the robot, including drag, added mass, and buoyancy. These forces are dynamically modeled by treating the submerged length of the legs as time-dependent, allowing for continuous adjustments in the projected area and submerged volume. The resulting torque on the robot’s center of mass is calculated based on these time-varying quantities, reflecting the interaction between the robot and the surrounding fluid. This approach ensures accurate simulation of the fluid resistance encountered during wading, enabling realistic robot adaptation to varying water conditions.


\subsubsection{Deformable Terrain}
\begin{figure}[t]
  \centering
  \includegraphics[width=0.95\linewidth]{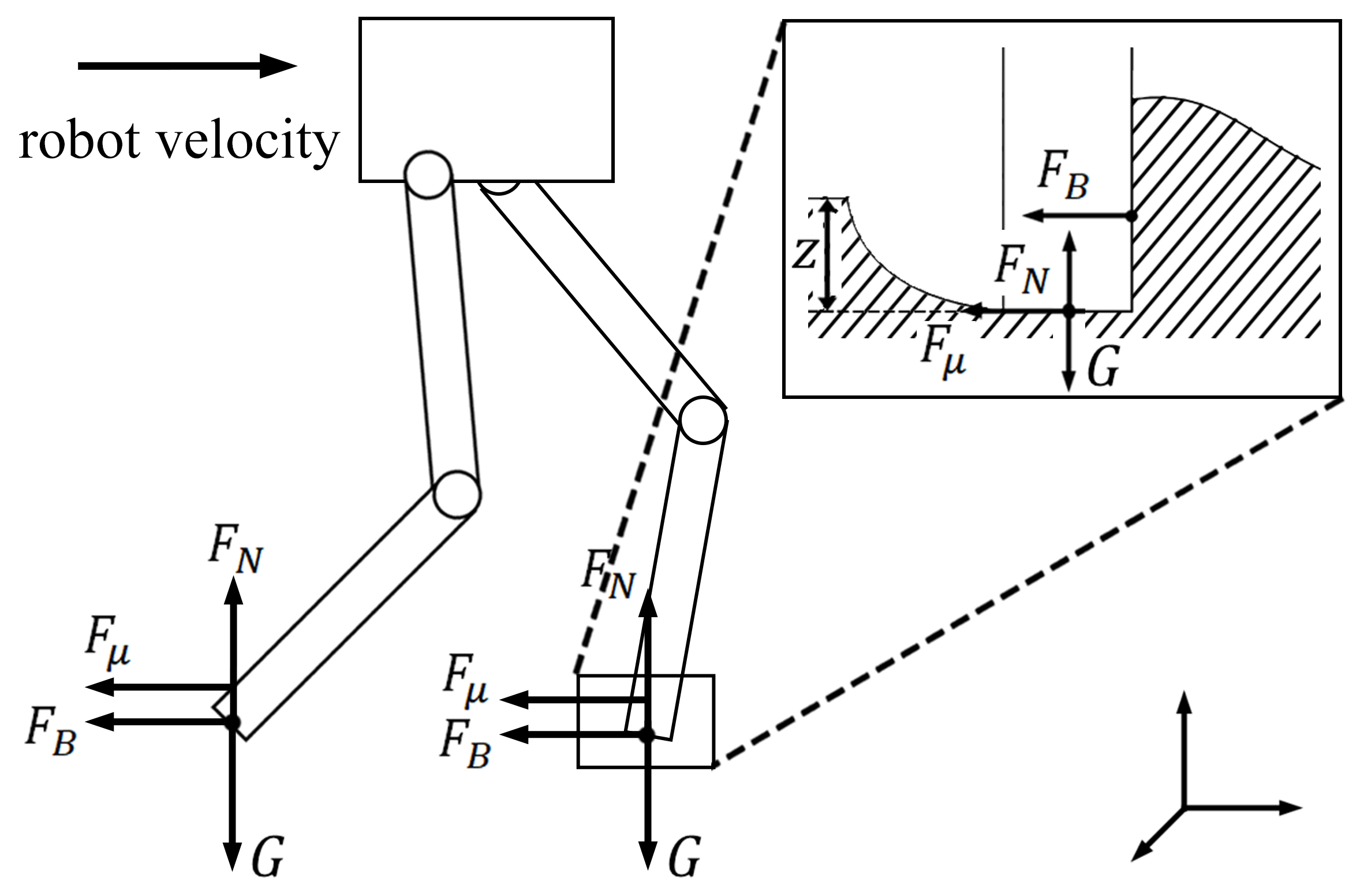}
  \caption{Analysis of bipedal robots walking on deformable terrains.}
  \label{fig:defo}
\end{figure}

Analyzing the forces exerted on a legged robot walking on deformable terrains, such as sand, requires a detailed understanding of both the mechanical properties of the terrain and the interaction between the robot's foot and the ground. We adopt a contact model following the work of Vanderkop et al. \cite{c21}, which identifies frictional forces and bulldozing resistance as the dominant forces accurately predicting the shear forces between the terrain and the foot, as illustrated in \cref{fig:defo}.

The total horizontal force acting on the robot's foot can be decomposed to the frictional force $F_{\mu}$ and the bulldozing resistance $F_B$. They are influenced by the physical properties of the terrain, including moisture content, grain size, cohesion, internal friction angle, and bulk density. The vertical force includes gravity $G$ and supporting force $F_N$.

The \textbf{bulldozing resistance} is caused by piled up soils against the sink foot. The direction of $F_B$ is opposite the direction of foot velocity, defined as: 
\begin{equation}
    F_B = az^n,
\end{equation}
where $z$ is the sinkage, both $a$ and $n$ are empirical parameters, named by bulldozing coefficient and bulldozing exponent, respectively.

\textbf{Frictional force} As displacement between two contacting surfaces increases, friction transitions from a static pre-sliding regime to a dynamic gross sliding regime~\cite{c25}, where it stabilizes at a steady-state value as long as relative velocity remains constant. This behavior is modeled using the Coulomb friction model, which distinguishes between static and kinetic friction phases, as defined in \cref{eq:Fmu}. It offers a simplified yet effective representation of frictional forces based on the normal load and differing static and dynamic friction coefficients.

\begin{equation}
    F_{\mu} = \mu_f F_N (1 - e^{-x/K}),
    \label{eq:Fmu}
\end{equation}
where $\mu_f$ is the coefficient of friction, $x$ is the displacement of the foot along the soil, and $K$ determines the rate of increase of friction in the pre-sliding regime. $F_N$ determines the steady state friction force in the gross sliding regime,

In our approach, parameters are selected based on previous experimental results and randomly sampled from a Gaussian distribution. This approach captures natural variability in terrain conditions and makes the model more adaptable to diverse environments. The randomness helps avoid overfitting, ensuring the system remains generalizable across different scenarios and supports a more thorough evaluation of the system's performance by testing its resilience under a wide range of parameter configurations.


In summary, the horizontal external force simulation $\mathbf{F}_{x}$ can be summarized as an inner product:
\begin{equation}
    \mathbf{F}_{x} = < \begin{bmatrix}
            \xi\mathbf{1}_w \\
            \xi\mathbf{1}_w \\
            \xi\mathbf{1}_d \\
            \xi\mathbf{1}_d
            \end{bmatrix},
            \begin{bmatrix}
            F_d(t) \\
            F_\text{flow}(t) \\
            F_b \\
            F_{\mu} \\
            \end{bmatrix} >, 
\end{equation} 
where $\xi$ is a Gaussian noise, defined as $\xi \sim \mathcal{N}(1, 0.1)$, $\mathbf{1}_w$ and $\mathbf{1}_d$ are indicator functions that serve as flags to apply specific physical properties on the generated terrain.

\begin{figure*}[t]
  \centering
  \includegraphics[width=0.95\linewidth]{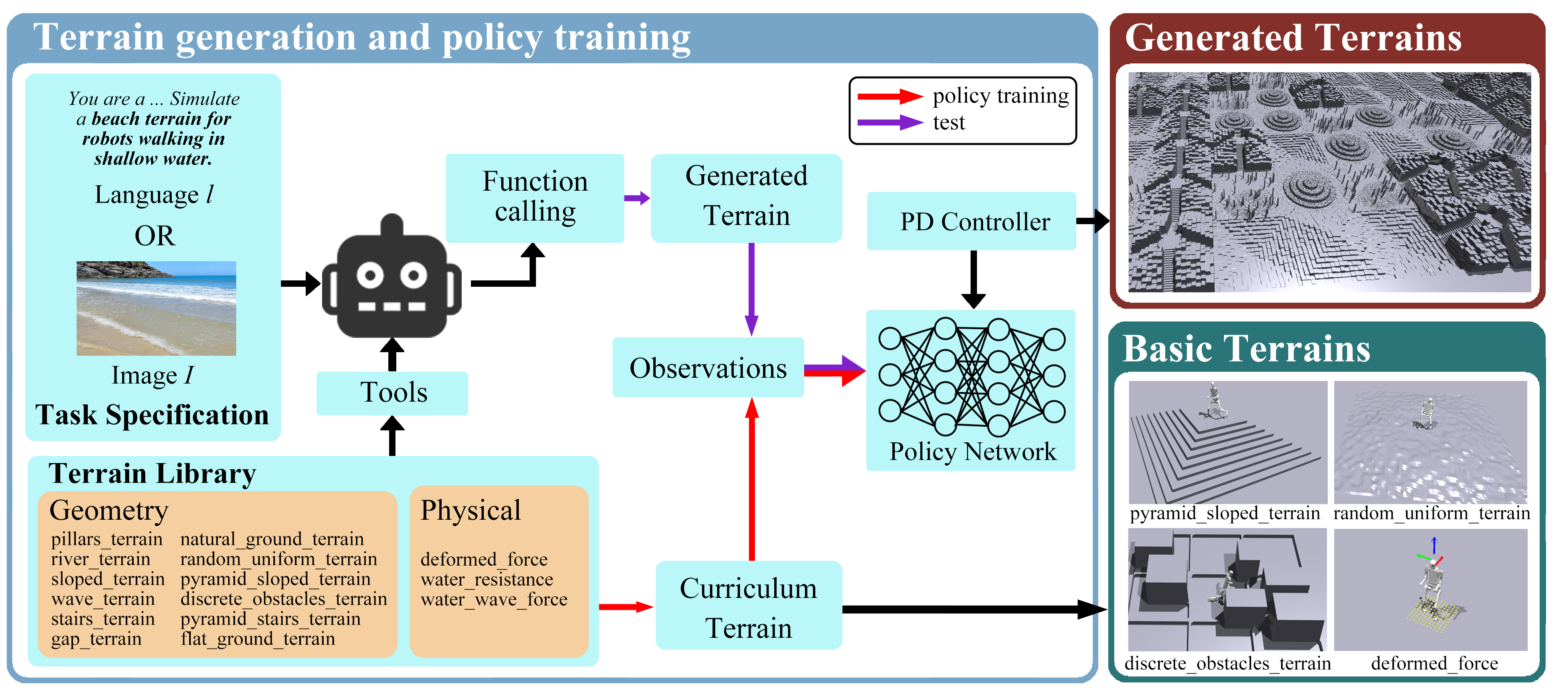}
  \caption{\textbf{Structure of GenTe}. The proposed pipeline for terrain generation and policy training in a simulated environment for bipedal robot locomotion control. During \textcolor{myred}{policy training}, the robots are trained on individual basic terrains in a curriculum-based progression. In the \textcolor{mypurple}{inference} phase, the task specification is provided either as a text description or an image. The LLM then calls functions from the terrain curriculum to apply relevant geometry and physical properties, creating a targeted, real-world-inspired terrain. Examples of basic terrains and generated terrains are shown on the right.}
  \label{fig:pipeline}
\end{figure*}

\subsection{Obstacles avoidance}
The \textbf{pillars terrain} and \textbf{rocks terrain} introduce a combination of obstacles to evaluate the robot’s obstacle avoidance capabilities. The pillars terrain simulates tall, high obstacles such as trees and street lamps, while the rocks terrain represents lower obstacles like rocks and garbage bins. During policy training, collisions with these obstacles are penalized, encouraging the robot to develop effective avoidance strategies. The robot learns to recognize the positions and orientations of obstacles, adjusting its path and movement to avoid collisions. This training ensures that the robot performs well in clear terrains and demonstrates the agility and decision-making required to navigate more complex, cluttered environments.

The diverse terrains developed in this simulation environment provide a comprehensive training ground for the bipedal robot. By tackling fluid resistance, varying levels of traction, deformable surfaces, and complex obstacle configurations, the robot can develop robust locomotion strategies. This diverse set of terrains ensures that the robot is equipped to handle real-world environments with complex, dynamic challenges.

\section{TERRAIN GENERATION FOR POLICY TRAINING}

In this study, we propose a novel approach to generate terrains for legged robot locomotion policy training. Our approach uses VLMs to automatically generate diverse terrains, which combines function-calling techniques with domain randomization to produce a wide range of terrain types for robust policy training. The pipeline of the proposed method is shown in \cref{fig:pipeline}. 

\subsection{VLM-Driven Terrain Generation in GenTe}

In the zero-shot inference phase, the GenTe framework leverages Vision-Language Models (VLMs) as a reasoning engine to transform high-level terrain descriptions—whether textual or visual—into simulator-ready environments with both geometric and physical realism. The process involves three key phases: semantic interpretation, function-based parameterization, and terrain construction.

In the first phase, semantic interpretation, the VLM interprets textual or visual input, extracting key features and relationships that define the terrain's nature. This allows the model to grasp not only the surface features of the terrain but also its underlying properties. During function-based parameterization, the interpreted features are mapped to a set of parameters that correspond to the terrain's geometry and physical properties, such as surface roughness, friction, and elevation gradients. Finally, in the terrain construction phase, these parameters are used to procedurally generate a simulator-ready environment, ensuring the terrain’s realism in terms of both geometry and physics, allowing for dynamic interactions in robotic simulations. This approach provides a flexible and scalable method for generating complex terrains based on minimal high-level input.

\subsection{Policy Training}

We model the terrain generation and control problem as a Discrete-time Markov Decision Process (DTMDP). In this context, we define a policy $\pi_{\theta}(a | s, m)$ as a probabilistic mapping from states $s \in \mathcal{S}$ to actions $a \in \mathcal{A}$. The goal of the policy $\pi_{\theta}$ is to maximize the expected cumulative reward over all tasks, which can be expressed as:
\begin{equation}
    \pi_{\theta}^* = \mathbb{E}_{\pi_{\theta}} \left[ \gamma^t \sum_{t=0}^{\infty} \mathcal{R}(s_t, a_t) \right],
\end{equation}
where $s_t$ and $a_t$ are the state and action at time step $t$, and $\gamma$ is the discount factor.

To execute the control commands generated by the policy, we employ a Proportional-Derivative (PD) controller, which translates high-level actions from the policy network into joint torques for the robot's actuators. Let $\tau$ denote the joint torques, $\theta_d$ the desired joint angles (derived from the policy output), and $\theta$ the current joint angles. The PD control law is given by:

\begin{equation}
    \tau = K_p (\theta_d - \theta) - K_d \dot{\theta},
\end{equation}
where $K_p$ and $K_d$ are the proportional and derivative gains, respectively, and $\dot{\theta}$ represents the joint velocity. This control setup ensures smooth, stable movement by continuously adjusting the torques to track the desired joint positions while damping oscillations.

\section{EXPERIMENTS AND RESULTS}

In this section, we present the experimental setup, training process, and evaluation results of our proposed method. During training, basic terrains were used for curriculum learning to help the agent acquire fundamental navigation skills. For evaluation, generated terrains were used. The experiments were conducted using the Proximal Policy Optimization (PPO) \cite{c22} for policy training, with key parameters including a learning rate of $5\times10^{-4}$ and a discount factor ($\gamma$) of 0.99. GenTe utilized Qwen2-VL-72B-Instruct \cite{c23} for terrain generation with image inputs, while Llama-3.1-8B-Instruct \cite{c24} was used for text input processing. The simulations were performed in Isaac Gym, where the generated terrains were structured as height maps with physical attributes, ensuring compatibility with various simulators.

\begin{figure}[t]
  \centering
  \includegraphics[width=1\linewidth]{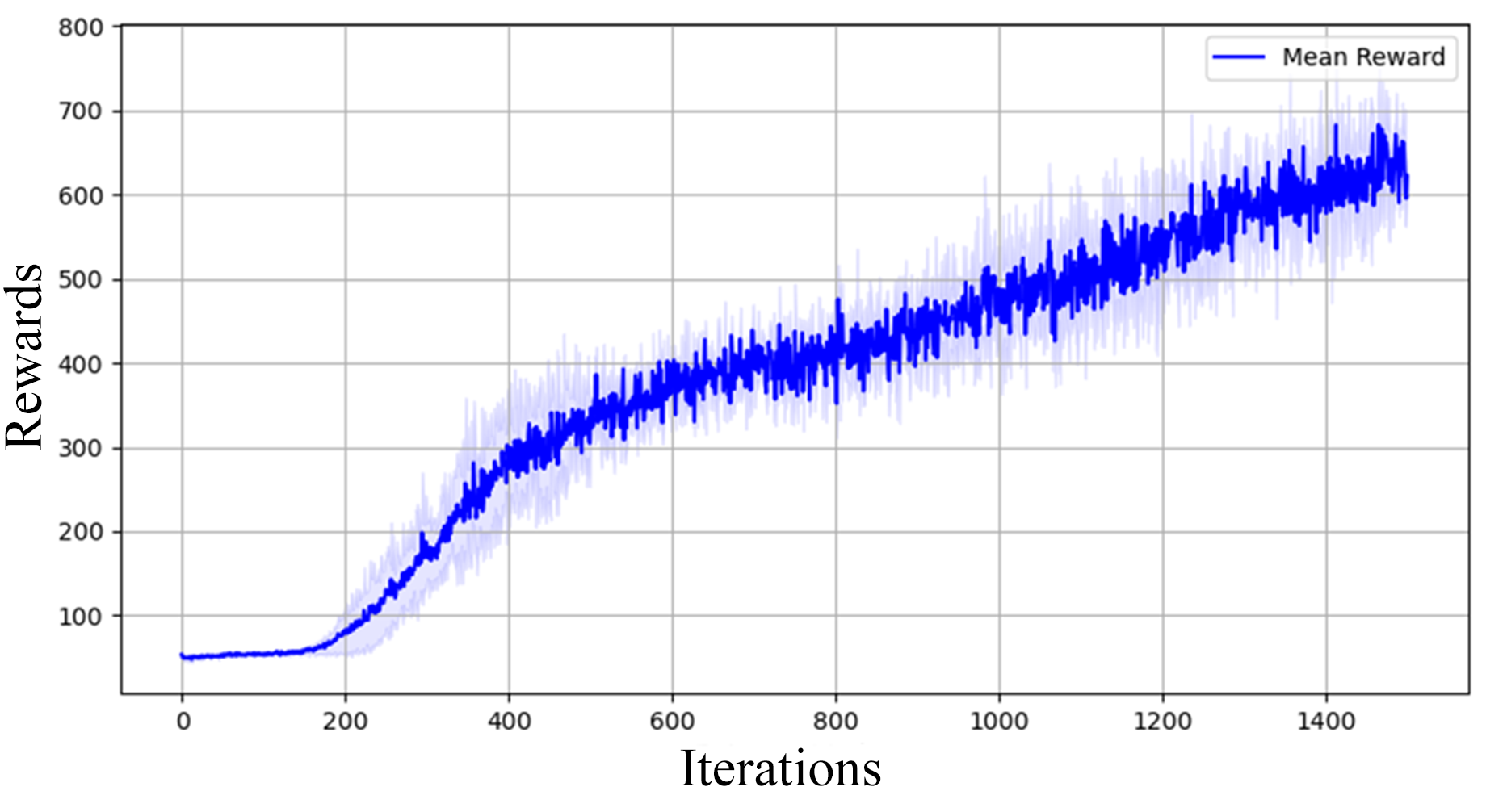}
  \caption{Averaged training reward with variance intervals over three trails. The solid blue line represents the mean reward over training iterations, while the shaded region indicates the variance, demonstrating the stability and improvement of the agent’s performance.}
  \label{fig:reward}
\end{figure}

\subsection{Policy Training}
\begin{figure}[t]
  \centering
  \includegraphics[width=1\linewidth]{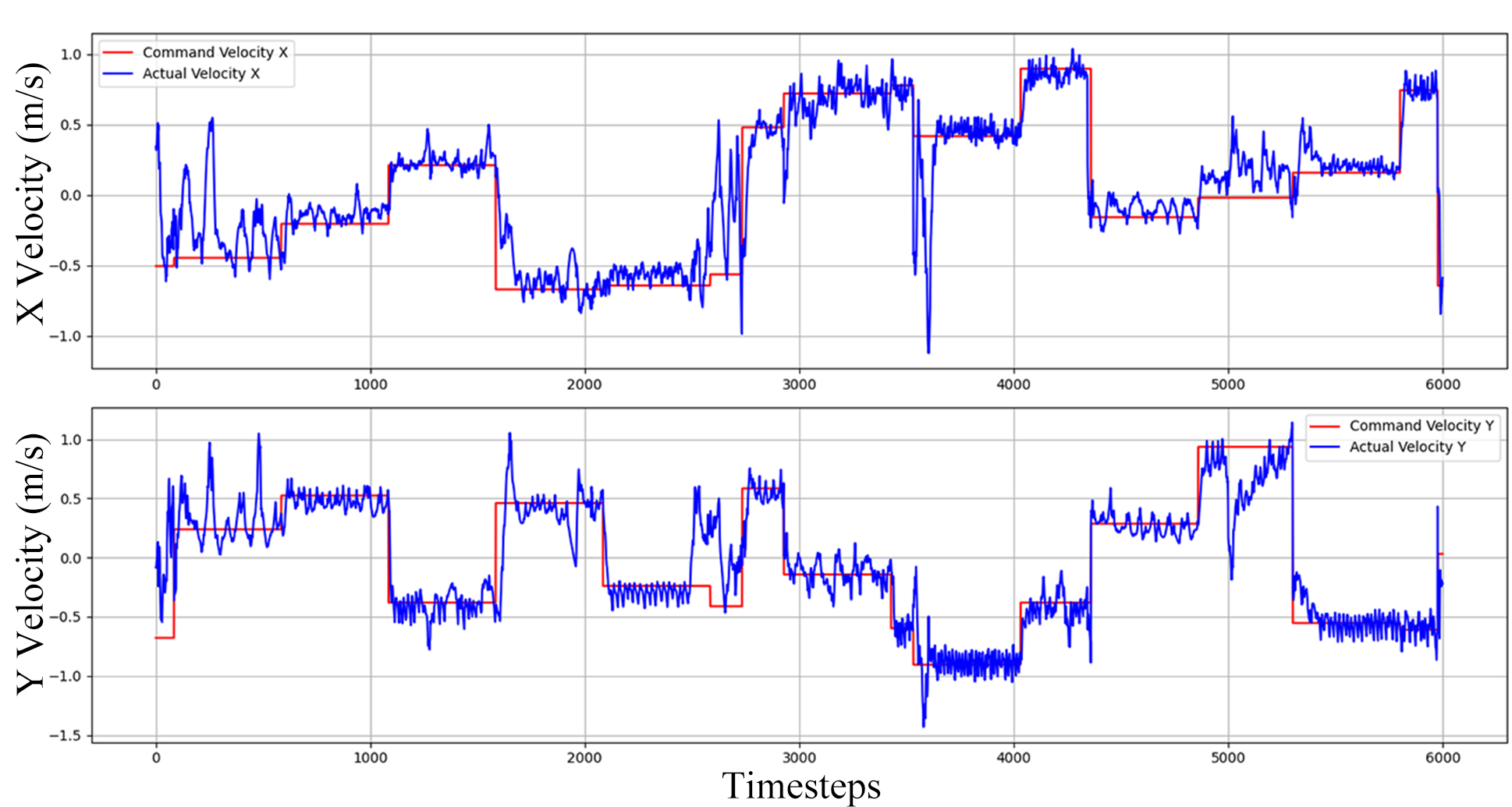}
  \caption{Command-following performance across different terrains of X-axis velocity (upper), and Y-axis velocity (lower). The red lines are the command velocity and the blue lines are actual velocity of robots.}
  \label{fig:cmd}
\end{figure}

The policy was trained for 1,500 iterations using curriculum learning on fundamental terrains. As depicted in \cref{fig:reward}, the cumulative reward over three trials exhibits a steady increase in mean reward, demonstrating effective learning and performance enhancement. The variance intervals further indicate the stability of the training process. To assess the agent’s command-following capability, we evaluated its response to velocity-based navigation instructions. The results, presented in \cref{fig:cmd}, highlight the policy’s ability to generalize to previously unseen terrains while maintaining stable control and accurate execution of the given commands.

\begin{figure}[h]
  \centering
  \includegraphics[width=1\linewidth]{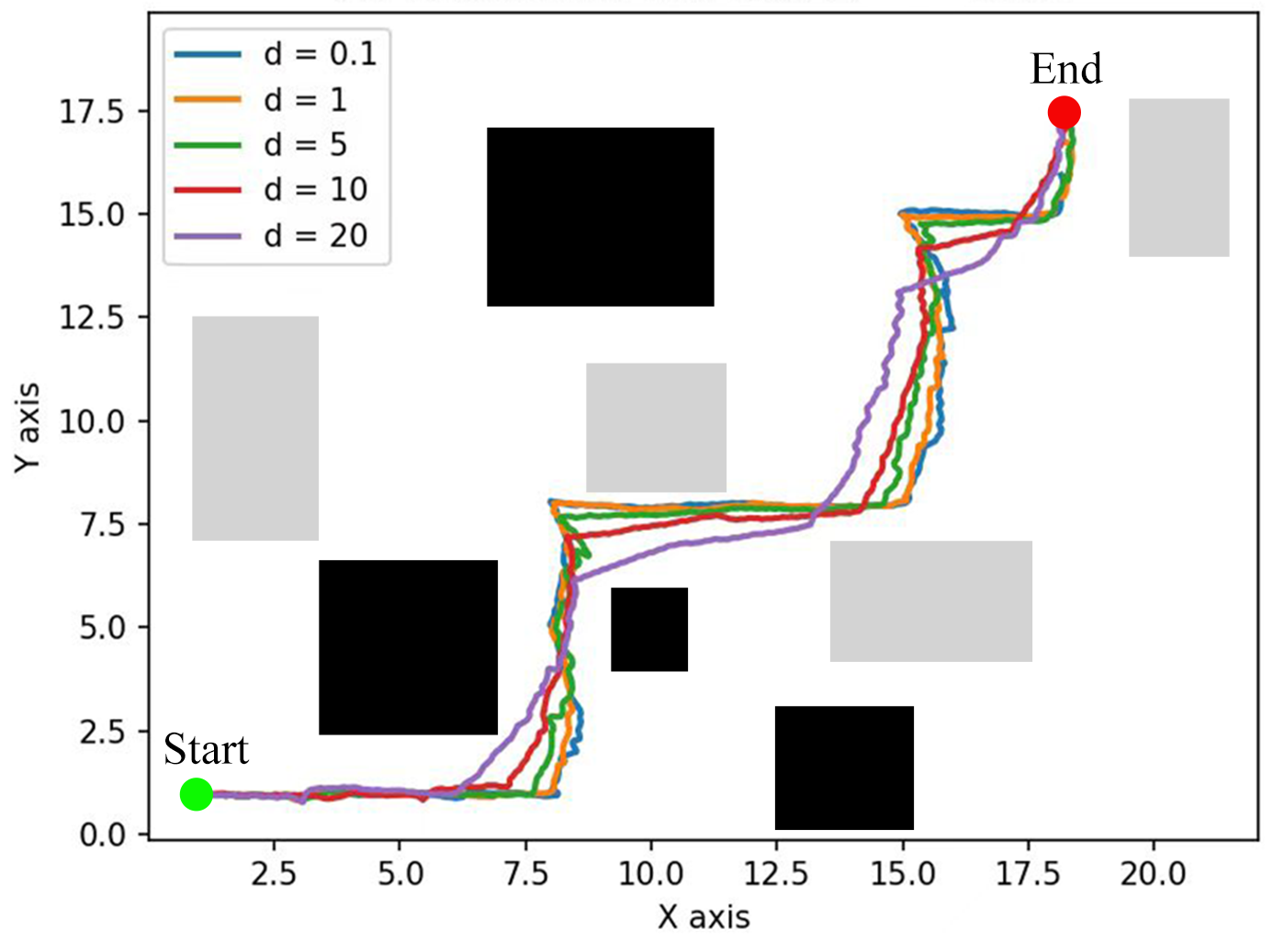}
  \caption{Overhead view of trajectory paths overlaid on a height map. The gray scale representation indicates terrain elevation, with black representing higher altitudes and white indicating lower ones. Robots receive directional commands that update when they approach a pre-defined way-point within a distance $d$. $d$ varies from 0.1 meters to 20 meters. }
  \label{fig:obs}
\end{figure}

\begin{figure*}[t]
  \centering
  \includegraphics[width=1\linewidth]{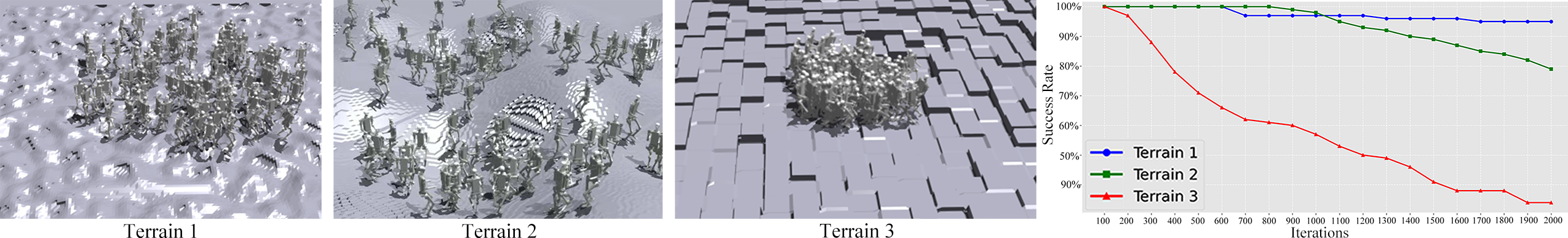}
  \caption{Success rate over iterations in three generated terrains (left). The terrains are shown on the right. The graph demonstrates the agent’s ability to adapt to novel environments while maintaining stable locomotion. Robots started at the center of the simulation environment and follow random direction commands.}
  \label{fig:sr}
\end{figure*}

\subsection{Obstacle Avoidance Capability}

The agent's ability to autonomously avoid obstacles was evaluated in a terrain with varying obstacles. As depicted in \cref{fig:obs}, black and gray blocks indicates obstacle with different height. The robots begin at the green "Start" point and navigate towards the red "End" point while avoiding obstacles. As $d$ increases, the trajectories exhibit smoother and more direct paths, suggesting that larger distance thresholds lead to fewer course corrections and potentially more efficient navigation. Results demonstrate that the trained policy effectively navigates around obstacles while adhering to designated way-points. The comparison of different $d$ values provides insight into the effect of way-point update frequency on trajectory smoothness.

\subsection{Evaluation on Generated Terrains}

To comprehensively evaluate the robustness of our proposed method, we deployed the trained policy in a variety of previously unseen terrains generated by the VLM. The success rates across different terrain types, along with the corresponding velocity metrics, are depicted in \cref{fig:sr}. To further challenge the policy’s adaptability and stability, we conducted additional tests beyond the initial training phase, running the policy for more than 2,000 iterations— exceeding the number of iterations used during training. The results demonstrate that the policy consistently maintains high success rates even when encountering complex and unstructured environments, underscoring its strong generalization capability. This sustained performance across diverse and unpredictable terrains highlights the effectiveness of our terrain generation strategy in producing meaningful variations and the robustness of our training framework in fostering adaptive and resilient navigation. By demonstrating stable control, precise execution of commands, and reliable decision-making across a broad range of conditions, the trained policy proves its potential for real-world deployment in challenging and dynamic environments.

\section{CONCLUSION}
In this work, we introduced GenTe, a terrain generation framework that enhances the generalization capabilities of bipedal robot locomotion by constructing diverse and realistic training environments. By integrating geometric and physical terrain modeling, along with function-calling techniques in LLMs, GenTe enables scalable and adaptive terrain creation based on textual or graphical input. Experimental results validate the effectiveness of the generated terrains. The open-source release of our framework and benchmark aims to accelerate future research in legged robot learning and terrain-aware locomotion. Future work will explore real-world deployment and further refine physical terrain modeling for enhanced realism.

\addtolength{\textheight}{-12cm}   


\end{document}